\ifdefined\pdfobjcompresslevel

\fi

\documentclass[letterpaper,10pt,conference]{ieeeconf}

\IEEEoverridecommandlockouts
\usepackage[T1]{fontenc}
\usepackage{amsmath,amssymb,bm}
\usepackage{graphicx}
\usepackage{booktabs}
\usepackage{cite}
\usepackage{xcolor}
\usepackage{microtype}
\usepackage{url}
\usepackage{balance}
\usepackage{stfloats}

\newcommand{\bfp}{\mathbf{p}}
\newcommand{\bfv}{\mathbf{v}}

\newcommand{\bfa}{\mathbf{a}}
\newcommand{\bfq}{\mathbf{q}}
\newcommand{\bfn}{\mathbf{n}}
\newcommand{\bfomega}{\boldsymbol{\omega}}

\newcommand{\Ftgt}{F^{\mathrm{target}}}
\newcommand{\Fmeas}{F^{\mathrm{meas}}}
\newcommand{\vcmd}{v^{\mathrm{cmd}}}
\newcommand{\must}{\mu_{\mathrm{st}}}
\newcommand{\mukin}{\mu_{\mathrm{kin}}}
\newcommand{\vtan}{\mathbf{v}^{\mathrm{tan}*}}
\newcommand{\Ffric}{\mathbf{F}^{\mathrm{fric}}}
\newcommand{\vres}{v^{\mathrm{res}}}
\title{\LARGE \bf
Robotic Multiphase Interaction: Manipulating Coupled Liquid and Solid Dynamics with a World Model}

\author{
Yixuan Feng$^{1}$,
Peng Wang$^{2,*}$\\[0.4em]
\small
$^{1}$College of Design and Engineering, National University of Singapore, 9 Engineering Dr 1, Singapore,117575.\\
$^{2}$Centre for Vision, Speech and Signal Processing (CVSSP), University of Surrey, Guildford GU2 7XH, United Kingdom.\\
$^{*}$Corresponding author: peng.wang@surrey.ac.uk
}

\begin{document}
\maketitle
\thispagestyle{empty}
\pagestyle{empty}

\begin{abstract}
This work presents \emph{Robotic Multiphase Interaction (RMI)}, a setting in which liquid enters a porous material and interacts mechanically with its deforming solid skeleton. Manipulation can therefore change pore volume, expel or redistribute retained liquid, and alter grasp stability at the same time. Spilled liquid can also create safety risks in domestic and manufacturing settings. This differs from most manipulation of solid objects and from tasks that involve both liquid and solid while keeping the phases spatially separate. We study a sponge filled with water as the first RMI example. We use implicit incompressible porous flow with smoothed particle hydrodynamics as the dynamics engine and enable robotic manipulation by adding Coulomb contact memory, hybrid velocity and force regulation, and a stability gate for lifting. The resulting environment connects robot commands to changes in the coupled liquid and solid state. A world model conditioned on actions predicts how this state evolves under candidate commands, while a temporal UNet generates actions using either Diffusion Policy  or rectified flow matching. Our world model reduces retained water prediction error by more than $60\%$ compared with the baseline. The best action sequence selected by the world model from policy proposals further reduces the predicted terminal water error by about half. These improvements show that modelling the coupled liquid and solid state helps the robot predict how its actions affect both the porous object and the liquid held inside.

\end{abstract}

\section{Introduction}
\label{sec:intro}

Robotic manipulation research mainly targets solid objects, whether they are rigid or deformable. For rigid objects, the central variables are pose and contact. Methods for deformable objects additionally model shape, strain, or material response, but the manipulated object is still treated as a solid. These formulations describe how an action moves or deforms the object~\cite{mahler2017dexnet,sanchez2020gns,shi2022robocraft}.

Liquid manipulation has usually been studied as bulk transfer between containers or as interaction between a solid object and surrounding liquid. In these settings both phases may be present, but they occupy different regions: the liquid remains outside the solid~\cite{li2019dpi}. A porous object filled with liquid breaks this separation. The liquid occupies the solid's pore space, exchanges momentum with the skeleton, and moves as that skeleton deforms. Fig. 1 shows these three manipulation settings. For a sponge filled with water, compression changes pore volume and pressure, expels or redistributes liquid, and alters contact loading and grasp stability. Neither the solid state nor the liquid state alone is sufficient to describe the manipulation outcome.

\begin{figure}[t]
\centering
\includegraphics[width=\columnwidth]{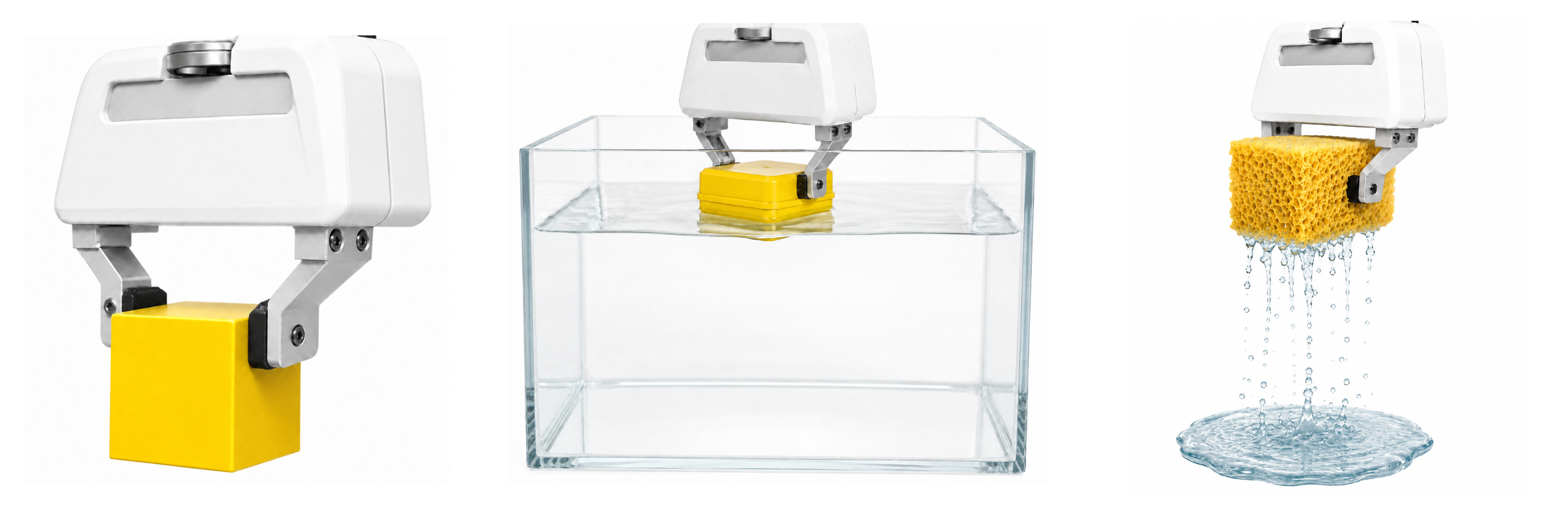}
\caption{Three manipulation settings. \textit{Left}: conventional manipulation of a solid object, which may be rigid or deformable. \textit{Middle}: both a solid and liquid are present but occupy different regions. \textit{Right}: porous RMI, where liquid occupies the solid's pore space and interacts with its deforming skeleton.}
\label{fig:nailpic}
\end{figure}

We frame this problem as \emph{Robotic Multiphase Interaction (RMI)}. An RMI task has two or more material phases that occupy the same region and interact mechanically, robot actions that change their coupling or transport, and an objective defined on the resulting phase state.
\begin{subequations}\label{eq:rmi}
\begin{align}
&\mathbf s_{t+1}^{\rm phase} = \mathcal F_{\rm RMI}(\mathbf s_t^{\rm phase},\mathbf u_t), \label{eq:rmi_state}\\
&\mathbf q_{t+1}^{\rm task} = \mathcal G(\mathbf s_{t+1}^{\rm phase}), \label{eq:rmi_task}
\end{align}
\end{subequations}
where $\mathbf s_t^{\rm phase}$ is the coupled material state, $\mathbf u_t$ is the robot action, and $\mathcal F_{\rm RMI}$ describes how the action changes the phases. The task quantity $\mathbf q_{t+1}^{\rm task}$ is obtained from the evolved phase state through the mapping $\mathcal G$. This definition distinguishes RMI from manipulation that merely involves several materials. A sealed bottle or a solid immersed in liquid is not an RMI task simply because both phases are present. The robot action must alter their coupled evolution, and that change must matter to the task.

RMI covers a broad range of tasks, and this work studies porous liquid and solid manipulation as its first instance. We use Implicit Incompressible Porous Flow with Smoothed Particle Hydrodynamics (SPH)~\cite{boettcher2025porous} as the dynamics engine. It represents overlapping liquid and porous solid sampling domains and couples pressure, drag, capillarity, buoyancy, viscosity, and solid elasticity. SPH provides the internal phase physics, but it does not provide stable gripper contact, robot force regulation, a safe transition to lifting, or a compact prediction of the coupled state under an action.

We bridge this gap by adding a robot interaction layer to SPH. Coulomb friction with stored contact points prevents drift during holding and lifting. A hybrid controller closes the grippers at a prescribed velocity until contact and then regulates $\Ftgt$. Lifting begins only after bilateral contact loads and sponge motion have stabilised. For efficiency, the robot does not observe the full particle state $\mathbf s_t^{\mathrm{phase}}$ in Eqs.~\eqref{eq:rmi_state} and~\eqref{eq:rmi_task}. Instead, it observes a compact interaction state $\mathbf y_t$ that combines liquid mass, saturation, and pore volume with the loading, contact, and slip through which $\mathbf u_t$ affects them. This interaction layer turns SPH into a platform for collecting trajectories used to train the world model and action policy.

As summarised in Fig. 2, we learn a compact world model that predicts how robot actions change the liquid and solid state, while a goal-conditioned policy proposes commands for a desired retained liquid outcome. Experiments compare both components with trajectories collected from the robot augmented SPH environment. Our contributions are threefold: (1) we define RMI as a manipulation problem in which robot actions change the coupling between material phases that occupy the same region, including liquid inside a porous solid; (2) we develop a robot interaction layer with Coulomb contact memory, force and velocity regulation, a stability gate for lifting, and an observation $\mathbf y_t$ that represents phase, load, contact, and slip; and (3) we develop a world model conditioned on actions, $W(\mathbf Y_t,\mathbf A_t)$, and a policy conditioned on a goal, $\pi(\mathbf A_t^{\pi}\mid\mathbf Y_t,\mathbf g_t)$, that proposes actions for desired liquid and solid interaction states.

\begin{figure*}[t]
\centering
\includegraphics[width=1.0\textwidth]{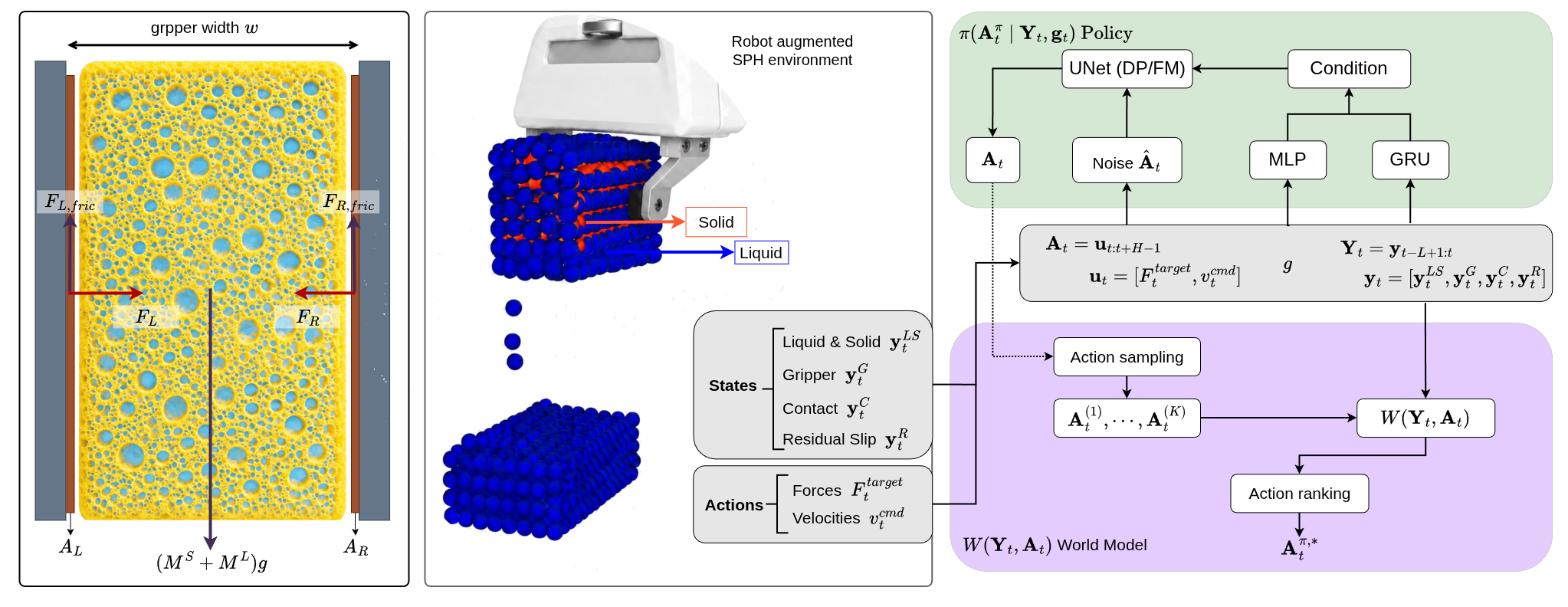}
\caption{RMI framework. \textit{Left}: two grippers grasp a sponge soaked with liquid, with the relevant forces, contacts, and gripper width. $M^S$ and $M^L$ denote the solid and retained liquid masses, respectively. \textit{Middle}: the state and action representations used in this work. \textit{Right}: the training and inference pipeline for the world models and policies.}
\label{fig:pipeline}
\end{figure*}

\section{Related Work}
\label{sec:related}

\textbf{Solid object manipulation.} Rigid grasping is commonly formulated around object pose and contact~\cite{mahler2017dexnet}. Learned dynamics extend manipulation to deformable and elastoplastic objects~\cite{sanchez2020gns,shi2022robocraft,zhang2024adaptigraph,huang2025particleformer}. Although these objects may bend, stretch, or permanently deform, the manipulated body is still represented primarily as a solid whose geometry or material state changes under action.

\textbf{Liquid and solid manipulation.} Pouring and dispensing methods regulate bulk liquid transfer~\cite{schenck2017pouring,huang2024kitchen}, while particle models can represent rigid, deformable, fluid, and multiple material phases~\cite{li2019dpi}. Wiping and liquid dabbing use a soft tool against a surface that carries liquid~\cite{tsuji2024wiping,hristov2021sponge}. These studies may involve both liquid and solid, but their robot objectives concern bulk transfer, external liquid motion, or a surface film. They do not represent the liquid as an internal phase occupying the deforming solid's pore space. Porous RMI instead predicts how a robot command jointly changes the solid skeleton, retained liquid, saturation, pore volume, and grasp stability.

\textbf{Porous multiphase simulation.} Macroscopic porous SPH introduced flow based on Darcy's law and bidirectional coupling between fluid and solid~\cite{lenaerts2008porous}, while implicit incompressible SPH enabled stable pressure projection at larger time steps~\cite{ihmsen2014iisph}. Implicit incompressible porous flow using SPH~\cite{boettcher2025porous} subsequently coupled overlapping liquid and porous solid domains through density correction for porosity, drag, capillarity, buoyancy, and elasticity. It provides the required liquid and solid dynamics but not a robot interface for acting on and observing the coupled phases. We therefore extend it with stable grasping, hybrid control, an observable interaction state, and prediction conditioned on actions.

\section{Porous RMI and Robot Interaction}
\label{sec:method}

\subsection{Implicit Incompressible Porous Flow Physics}

We build the robot manipulation environment on implicit incompressible porous flow using SPH~\cite{boettcher2025porous}. Liquid and porous solid particles occupy overlapping sampling domains while remaining distinct phases. Their momentum exchange is
\begin{equation}
\mathbf f^{\mathrm{pore}}
=\mathbf f^{\mathrm{drag}}+\mathbf f^{\mathrm{cap}}
+\mathbf f^{\mathrm{buo}},
\label{eq:pore_force}
\end{equation}
where $\mathbf f^{\mathrm{drag}}$ opposes relative motion between the fluid and solid, $\mathbf f^{\mathrm{cap}}$ models capillary retention, and $\mathbf f^{\mathrm{buo}}$ is the force exerted on the solid skeleton by pressure.
Local saturation around solid particle $s$ is
\begin{equation}
S_s=\frac{1}{\phi}
\frac{\sum_{j\in\mathcal N_s^L}V_j^0\mathcal K_{sj}}
{\sum_{r\in\mathcal N_s^S}(m_r/\rho_r)\mathcal K_{sr}}\in[0,1],
\label{eq:saturation}
\end{equation}
where $\phi\in(0,1)$ is undeformed porosity, $\mathcal N_s^L$ and $\mathcal N_s^S$ are the liquid and solid SPH neighbours of $s$, $V_j^0$ is the rest sampling volume of liquid particle $j$, and $m_r$ and $\rho_r$ are the mass and current density of solid particle $r$, so $m_r/\rho_r$ is its current volume. The kernel term is $\mathcal K_{sa}=\mathcal K(\|\mathbf x_s-\mathbf x_a\|,h)$ for a neighbouring particle $a$. The numerator estimates local liquid content, while the denominator estimates the current volume of the porous solid.

Equations~\eqref{eq:pore_force} and~\eqref{eq:saturation} describe the physical loop behind RMI. Gripper compression changes pore volume and pressure, which drives liquid transport. The resulting saturation and retained mass then affect the material response and the load on the gripper. 

\subsection{Robot Interaction}
\label{sec:contact}

The incompressible porous flow with SPH does not specify stable gripper contact or robot actions. We add three components to make the SPH environment robot-interactive; together, they form the interaction layer. 
The interaction layer runs at the simulator time step, while the learning models use index $t$ after temporal resampling.

\paragraph{Coulomb friction with contact memory}
Memoryless tangential damping allows integration error to accumulate as the gripper holds and lifts the sponge. Let $s$ be a solid particle in contact with a gripper, $N_s\geq 0$ the magnitude of its normal contact force, $m_s$ its mass, $\Delta t$ the simulator time step, and $(\must,\mukin)$ the static and kinetic friction coefficients. The static and kinetic Coulomb bounds are $\must N_s$ and $\mukin N_s$, respectively. Each contacting particle stores an \emph{anchor}, which is its contact location in the gripper frame. The contact memory mechanism is illustrated in Fig. 3. Let $\mathbf e_s$ be the tangential displacement from that anchor and $\vtan_s$ the relative tangential velocity before friction is applied. The restoring demand and the force actually applied are
\begin{subequations}\label{eq:friction}
\begin{align}
&\mathbf F_{\mathrm{req},s} = \frac{m_s}{\Delta t}\left(\frac{\beta}{\Delta t}\mathbf e_s-\vtan_s\right), \label{eq:friction_demand}\\
&\kappa_s = \min\left(\mukin N_s,\frac{m_s\|\vtan_s\|}{\Delta t}\right), \label{eq:friction_limit}\\
&\mathbf F_{\mathrm{kin},s} = -\kappa_s\frac{\vtan_s}{\max(\|\vtan_s\|,\varepsilon_v)}, \label{eq:friction_kinetic}\\
&\mathbf F_{f,s} = \begin{cases}
\mathbf F_{\mathrm{req},s}, & \|\mathbf F_{\mathrm{req},s}\|\leq\must N_s,\\
\mathbf F_{\mathrm{kin},s}, & \|\mathbf F_{\mathrm{req},s}\|>\must N_s.
\end{cases} \label{eq:friction_applied}
\end{align}
\end{subequations}
Here $\beta\in(0,1]$ is the positional recovery gain and $\varepsilon_v>0$ prevents division by zero. Equation~\eqref{eq:friction_limit} ensures that the kinetic correction cannot reverse the direction of slip within one integration step. The first condition in Eq.~\eqref{eq:friction_applied} defines sticking and retains the anchor, while the second defines sliding and releases it. Contact memory gradually removes accumulated drift without exceeding the Coulomb bound. The learned state keeps the aggregates $\vtan$ and $\Ffric$ over the contact set rather than storing every particle anchor. 

\begin{figure}[t]
\centering
\includegraphics[width=0.45\textwidth]{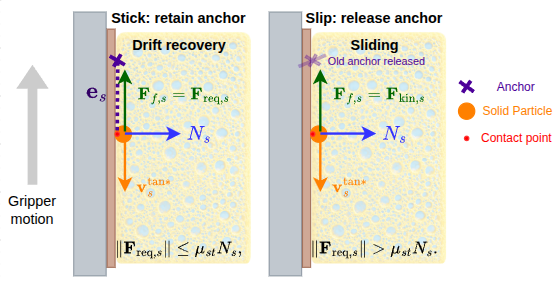}
\caption{Coulomb friction with contact memory. Each contacting solid particle stores an anchor in the gripper frame. The tangential displacement $\mathbf e_s$ and restoring demand $\mathbf F_{\mathrm{req},s}$ determine whether the contact sticks or slips within the Coulomb cone.}
\label{fig:forces}
\end{figure}

\paragraph{Hybrid velocity-force regulation}
Before contact, both grippers close at a prescribed speed. After the normal force remains above its threshold for $K_{\mathrm{contact}}$ steps, the controller switches to force regulation:
\begin{subequations}\label{eq:action}
\begin{align}
&\mathbf u_t = [\Ftgt_t,\vcmd_t], \label{eq:action_vector}\\
&\vcmd_t = \operatorname{clip}\bigl(k_F(\Ftgt_t-\Fmeas_t),-v_{\mathrm{open}},v_{\mathrm{close}}\bigr). \label{eq:force_regulation}
\end{align}
\end{subequations}
Here $\vcmd_t$ is the gripper velocity generated from the force error. A positive value closes the grippers when the measured load is too small, while a negative value opens them after an overshoot. The target $\Ftgt_t$ ramps smoothly from the force measured at contact, and guards on maximum force and minimum gripper width remain active. This controller realises a supplied load; it does not choose the force that is best for liquid removal. The variables $\Ftgt$ and $\vcmd$ are not included in the observation $\mathbf y_t$, so the forecast cannot access the command except through its action input.

\paragraph{Stability gate for lifting}
Reaching $\Ftgt$ alone does not make a grasp safe to lift. Let $\mathcal C_t$ be the set of solid particles touching either gripper, $\tilde F_{L,t}$ and $\tilde F_{R,t}$ the filtered gripper loads, and $v_{\mathrm{COM},t}$ the speed of the solid's centre of mass along the lift axis. The residual slip $\vres_t=|\mathcal C_t|^{-1}\sum_{s\in\mathcal C_t}\|\mathbf v_{s,t}^{\mathrm{tan,corr}}\|$ is the mean corrected tangential speed after $\mathbf F_{f,s}$ is applied. Lifting is enabled only when valid bilateral contact exists and
\begin{subequations}\label{eq:lift_gate}
\begin{align}
&F_{\min} \leq \tilde F_{L,t},\tilde F_{R,t}\leq F_{\max}, \label{eq:lift_force_gate}\\
&v_{\mathrm{COM},t} \leq v_{\mathrm{COM,th}}, \label{eq:lift_motion_gate}\\
&\vres_t \leq v_{\mathrm{res,th}}. \label{eq:lift_slip_gate}
\end{align}
\end{subequations}
All three conditions must hold for $K_{\mathrm{stable}}$ steps. The force condition requires a valid bilateral grasp, the centre of mass condition requires the sponge to settle, and the residual slip condition requires tangential motion at the grippers to remain small. Averaging the slip magnitudes prevents opposite motions from cancelling. The gate therefore distinguishes 'force reached' from 'grasp settled enough to lift'. Both grippers then receive the same displacement.

\section{RMI World Model}
\label{sec:world_model}

The SPH solver evolves the complete particle field, but running it for every candidate action is too expensive and the robot does not need every particle position. Instead, it needs to know how retained liquid, pore volume, load, contact, and residual slip will change under a proposed action chunk $\mathbf A_t$, given the recent interaction history $\mathbf Y_t$. As illustrated in Fig. 2, we denote this action-conditioned query by the world model $W(\mathbf Y_t,\mathbf A_t)$. The causal models process the commands in sequence, while a lookahead diagnostic sees the complete chunk. Their predictions allow candidate actions to be evaluated by phase and grasp outcomes before execution.

\subsection{Interaction State and Actions}

The robot interaction layer provides a compact observation with four groups:
\begin{subequations}\label{eq:interaction_state}
\begin{align}
&\mathbf y_t^{LS} = [M_t^{L},S_t,V_{\mathrm{pore},t},V_{\mathrm{sponge},t}], \label{eq:state_liquid_solid}\\
&\mathbf y_t^{G} = [w_t,\Fmeas_t,F_{L,t},F_{R,t},\alpha_t^{F}], \label{eq:state_gripper}\\
&\mathbf y_t^{C} = [A_{L,t},A_{R,t},c_t,\vtan_t,\Ffric_t,\mathbf r_t^{\mathrm{fric}}], \label{eq:state_contact}\\
&\mathbf y_t^{R} = [\vres_t], \label{eq:state_stability}\\
&\mathbf y_t = [\mathbf y_t^{LS},\mathbf y_t^{G},\mathbf y_t^{C},\mathbf y_t^{R}], \label{eq:state_complete}
\end{align}
\end{subequations}
where $LS$ denotes the liquid and solid phase variables, $G$ the gripper variables, $C$ the contact variables, and $R$ the residual slip. In~\eqref{eq:state_liquid_solid}, $M_t^{L}$ is retained liquid mass, $S_t$ is saturation, $V_{\mathrm{pore},t}$ is current pore volume, and $V_{\mathrm{sponge},t}$ is current sponge volume. In~\eqref{eq:state_gripper}, $w_t$ is gripper width, $\Fmeas_t$ is the realised total load, $F_{L,t}$ and $F_{R,t}$ are the left and right gripper loads, and $\alpha_t^{F}=|F_{L,t}-F_{R,t}|/(F_{L,t}+F_{R,t}+\varepsilon_F)$ measures their imbalance, with $\varepsilon_F>0$. In~\eqref{eq:state_contact}, $A_{L,t}$ and $A_{R,t}$ are contact areas, $c_t\in\{0,1\}$ marks a valid contact set, $\vtan_t$ is the tangential slip tendency before friction is applied, $\Ffric_t$ is the applied tangential force, and $\mathbf r_t^{\mathrm{fric}}$ encodes one of four contact regimes: $\mathrm{none}$, $\mathrm{stick}$, $\mathrm{slip}$, or $\mathrm{mixed}$. In~\eqref{eq:state_stability}, $\vres_t$ is residual slip after friction is applied.

The contact variables are aggregated over the same set $\mathcal C_t$. Similar normal loads can occur during quiet sticking or active sliding, and these conditions affect drainage and lifting differently. When $\mathcal C_t$ is empty, $A_L=A_R=0$ and $\Ffric=\mathbf 0$ are physical zeros, while the friction label is $\mathrm{none}$. The value $\vtan=\mathbf 0$ then represents an empty average rather than stationary contact, and $c_t$ distinguishes these cases. Residual slip, $\vres=|\mathcal C_t|^{-1}\sum_{s\in\mathcal C_t}(\cdot)$, is undefined when $\mathcal C_t=\varnothing$, so only this value is masked.

These four groups form the state consumed by the world model. Retained liquid mass $M^{L}$ is the task quantity, while saturation $S$, pore volume $V_{\mathrm{pore}}$, and sponge volume $V_{\mathrm{sponge}}$ distinguish liquid loss from compression of the porous skeleton. The same amount of liquid in a smaller pore volume is a different phase state. Gripper width and load describe how strongly the sponge is held, while $F_L$, $F_R$, and $\alpha^{F}$ show whether the load is shared evenly. Contact area, $\vtan$, $\Ffric$, the friction regime, and $\vres$ describe whether that load is supported by a stable grasp. Spatial saturation remains in the particle field; the world model uses the aggregates needed to predict how much liquid remains and whether lifting is safe.

The trajectories are divided into training windows drawn from the controller phases \textit{close}, \textit{ramp}, \textit{hold}, \textit{lift}, and \textit{squeeze}. These phase labels are not included in the state $\mathbf y_t$. Particle positions, internal controller variables, and centre of mass speed are also omitted. The robot action is $\mathbf u_t=[\Ftgt_t,\vcmd_t]$, with force in newtons and velocity in metres per second. At time $t$, the model receives
\begin{subequations}\label{eq:history_action}
\begin{align}
\mathbf Y_t &= \mathbf y_{t-L+1:t}, \label{eq:history_tensor}\\
\mathbf A_t &= \mathbf u_{t:t+H-1}, \label{eq:action_tensor}
\end{align}
\end{subequations}
where $\mathbf Y_t$ contains the previous $L$ observations and $\mathbf A_t$ contains the next $H$ commands. Actions remain separate from $\mathbf y_t$, so the state never contains the command that the model is asked to evaluate. The sampling interval and values of $L$ and $H$ are given in Sec.~\ref{sec:experiments}.

Continuous channels in $\mathbf y$ and $\mathbf u$ are standardised using the training partition, while $c_t$ and $\mathbf r_t^{\mathrm{fric}}$ remain categorical. Before standardisation, the data loader creates a validity mask $\mathbf m_t$ in the same order as $\mathbf y_t$. An entry $m_{t,j}$ is one when $y_{t,j}$ is finite and available and zero otherwise. The mask is computed for each channel rather than copied from the contact label $c_t$. Quantities with a physical zero are set to zero when there is no contact, and the friction category is set to $\mathrm{none}$. Missing standardised values are filled with zero, which is also the standardised training mean, so $\mathbf m_t$ tells the model whether that value is valid. The mask is preprocessing information, not a physical state or prediction target.

The residual heads predict the 18 continuous entries $\mathbf z_t$ of $\mathbf y_t$. Contact $c_t$ and the friction class $\mathbf r_t^{\mathrm{fric}}$ use separate prediction heads. Every world model returns continuous states $\hat{\mathbf Z}$, contact probabilities $\hat{\mathbf p}^{c}$, and friction probabilities $\hat{\mathbf P}^{\mathrm{fric}}$ over the horizon. The models differ only in their inputs and how those inputs are processed. Table I summarises the causal and lookahead predictors. Once a recurrent model predicts a step, its generated entries are treated as valid at the next step.

\subsection{Causal and Lookahead Predictors}
\label{sec:baselines}

The world model block in Fig. 2 is instantiated by predictors that consume the same planned action chunk $\mathbf A_t$ in two different ways. AC (Action Conditioned)-GRU (Gated Recurrent Unit)-WM (World Model) processes one command at a time, so decoder step $\tau$ receives $\mathbf u_{t+\tau-1}$ through an MLP encoder $E_u$. AC-MLP (Multilayer Perceptron)-WM receives the complete flattened chunk $\operatorname{vec}(\mathbf A_t)$, so even its first prediction can depend on later commands, including $\mathbf u_{t+H-1}$. AC-GRU-WM provides the causal prediction that the robot can query as actions unfold, whereas AC-MLP-WM tests how well a model can fit a trajectory when the complete plan is already known.

\begin{table}[t]
\caption{World model predictors. Causal models receive one command at each step, while lookahead models see the complete action chunk at once.}
\label{tab:baselines}
\centering
\scriptsize
\setlength{\tabcolsep}{3pt}
\begin{tabular}{clp{0.50\columnwidth}}
\toprule
Interface & Model & Role \\
\midrule
Causal & Baseline & The latest state through the rollout. \\
Causal & State-GRU-WM & Same GRU as AC-GRU-WM, no command. \\
Causal & AC-GRU-WM & Proposed world model, one command per step. \\
Lookahead & Linear-ARX & Locally linear residual on the full chunk. \\
Lookahead & AC-MLP-WM & One evaluation of $(\mathbf y_t,\operatorname{vec}(\mathbf A_t))$. \\
\bottomrule
\end{tabular}
\end{table}

\paragraph{Causal world model}
AC-GRU-WM uses a GRU with two layers and width 128 to encode $\mathbf Y_t$. It embeds each command with $E_u$ and predicts changes in the standardised continuous state. Its overall mapping is
\begin{equation}
(\hat{\mathbf Z}_{t+1:t+H},\hat{\mathbf p}^{c}_{t+1:t+H},
\hat{\mathbf P}^{\mathrm{fric}}_{t+1:t+H})
=W_{\mathrm{GRU},\theta}(\mathbf Y_t,\mathbf A_t).
\label{eq:wm_map}
\end{equation}
Here $\theta$ denotes the model parameters, while $\phi$ remains reserved for porosity. An overbar indicates that the continuous entries of a state or action have been standardised; categorical entries are unchanged. At each prediction step, the model updates
\begin{subequations}\label{eq:decoder}
\begin{align}
&\mathbf d_{t+\tau-1} = [\hat{\mathbf y}_{t+\tau-1},\hat{\mathbf m}_{t+\tau-1},E_u(\bar{\mathbf u}_{t+\tau-1})], \label{eq:decoder_input}\\
&\mathbf h_{t+\tau} = \operatorname{GRU}_{\mathrm{dec}}(\mathbf d_{t+\tau-1},\mathbf h_{t+\tau-1}), \label{eq:decoder_hidden}\\
&\hat{\bar{\mathbf z}}_{t+\tau} = \hat{\bar{\mathbf z}}_{t+\tau-1}+D_z(\mathbf h_{t+\tau}), \label{eq:decoder_state}
\end{align}
\end{subequations}
Equation~\eqref{eq:decoder_input} combines the previous prediction, its validity mask, and the current command embedding. At the first decoder step, the mask comes from the latest observation; entries generated later in the rollout are marked as valid. Equation~\eqref{eq:decoder_hidden} updates the hidden state, and $D_z$ in Eq.~\eqref{eq:decoder_state} predicts the change in the continuous state. The command therefore enters at every step rather than only at the start of the rollout. Separate heads predict contact and friction. Training minimises
\begin{subequations}\label{eq:loss}
\begin{align}
&\ell_{t+\tau,d} = \operatorname{Huber}(\hat{\bar z}_{t+\tau,d},\bar z_{t+\tau,d}), \label{eq:continuous_element_loss}\\
&\mathcal L_{\mathrm{cont}} = \sum_{\tau,d}\gamma^{\tau-1}m_{t+\tau,d}\ell_{t+\tau,d}, \label{eq:continuous_loss}\\
&\mathcal L_{\mathrm{WM}} = \mathcal L_{\mathrm{cont}}+\lambda_c\mathcal L_{\mathrm{BCE}}+\lambda_f\mathcal L_{\mathrm{CE}}. \label{eq:world_model_loss}
\end{align}
\end{subequations}
Equation~\eqref{eq:continuous_element_loss} gives the error for one continuous channel at one prediction step. In Eq.~\eqref{eq:continuous_loss}, $\gamma\in(0,1]$ gives slightly less weight to later steps and $m_{t+\tau,d}$ removes undefined targets, mainly $\vres$ before contact. The weights $\lambda_c$ and $\lambda_f$ control the contact and friction losses in Eq.~\eqref{eq:world_model_loss}. These classification losses prevent accurate water and force predictions from hiding an incorrect contact mode. Teacher forcing is gradually reduced during training. We select the checkpoint with the lowest validation water MAE at 64 steps, using a patience of eight epochs and a maximum of 40.

State-GRU-WM is the matched control that sees only the state. It keeps the same encoder, decoder, and output heads as AC-GRU-WM but removes $E_u$ at every prediction step. The Baseline carries the latest continuous state, contact indicator, and friction label across the horizon. It provides a useful lower bound because liquid continues to move during the ramp, hold, lift, and squeeze, although much more slowly than during the first clamp. AC-GRU-WM is informative only if it improves on both controls. We also shuffle future action chunks while preserving their overall distribution. A drop in performance then shows that the model uses the command sequence rather than only the recent state history.

\paragraph{Lookahead residual}
Linear-ARX (Linear Autoregressive Model with Exogenous Inputs) fits a ridge residual with regularisation $10^{-2}$ to flattened histories and action chunks. It uses only training windows with complete targets. This baseline tests whether water and force behave almost linearly over a short horizon. It is not expected to represent the discrete friction regime well.

AC-MLP-WM asks whether the latest masked state and the complete future action chunk are enough to predict the rollout without recurrence. Its input is
\begin{equation}
\mathbf x_t^{\mathrm{MLP}}
=[\bar{\mathbf y}_t,\mathbf m_t,\operatorname{vec}(\bar{\mathbf A}_t)].
\label{eq:acmlp_input}
\end{equation}
The input contains the latest standardised state, its validity mask, and the flattened action chunk. Two shared hidden layers, each with width 128, give
\begin{subequations}\label{eq:acmlp_hidden}
\begin{align}
&\mathbf h_1 = \operatorname{ReLU}(W_1\mathbf x_t^{\mathrm{MLP}}+\mathbf b_1), \label{eq:acmlp_hidden_one}\\
&\mathbf h_2 = \operatorname{ReLU}(W_2\mathbf h_1+\mathbf b_2), \label{eq:acmlp_hidden_two}
\end{align}
\end{subequations}
Three output heads then produce
\begin{subequations}\label{eq:acmlp_heads}
\begin{align}
&\Delta\bar{\mathbf Z} = \operatorname{reshape}_{H\times 18}(W_z\mathbf h_2+\mathbf b_z), \label{eq:acmlp_continuous_head}\\
&\hat{\bar{\mathbf z}}_{t+\tau} = \bar{\mathbf z}_t+\sum_{k=1}^{\tau}\Delta\bar{\mathbf z}_k, \label{eq:acmlp_trajectory}\\
&\hat{\mathbf p}^{c} = \sigma(W_c\mathbf h_2+\mathbf b_c), \label{eq:acmlp_contact_head}\\
&\hat{\mathbf P}^{\mathrm{fric}} = \operatorname{softmax}_4\!\left(\operatorname{reshape}_{H\times 4}(W_f\mathbf h_2+\mathbf b_f)\right). \label{eq:acmlp_friction_head}
\end{align}
\end{subequations}
Equation~\eqref{eq:acmlp_continuous_head} predicts changes in the continuous state, and Eq.~\eqref{eq:acmlp_trajectory} accumulates them from the latest observation. The sigmoid $\sigma$ gives the contact probability, while $\operatorname{softmax}_4$ gives the four friction probabilities. We convert the continuous trajectory back to physical units before evaluation. Because the complete action chunk is visible when every step is predicted, lower water error here does not imply a better causal model than AC-GRU-WM.

\subsection{Policy and Ranking}
\label{sec:policy}

The policy is conditioned on the goal vector $\mathbf g_t=[g_t,\Delta M_t]$ and proposes $\mathbf A_t^{\pi}=\mathbf u_{t:t+H_\pi-1}$ through $\pi(\mathbf A_t^{\pi}\mid\mathbf Y_t,\mathbf g_t)$, with $H_\pi=32$. Here $g_t$ is the desired terminal retained liquid mass and $\Delta M_t=g_t-M_t^L$ is its change from the current value. During training, $g_t=M_{t+H_\pi}^{L}$ is taken from the terminal value in the collected trajectory. Both training objectives use the same temporal UNet, following~\cite{chi2023diffusion}. A GRU summarises the state history and validity mask, while an MLP encodes $\mathbf g_t$. FiLM passes this information to each residual block. The UNet has channel widths of 64, 128, and 256 and a kernel size of 5. It is trained with AdamW, a learning rate of $10^{-4}$, a batch size of 64, an exponential moving average of $0.999$, at most 80 epochs, and patience 12. Force commands are limited to $400\,\mathrm{N}$ for safety. All models are trained on an NVIDIA A100 GPU.

Diffusion Policy predicts the clean action chunk $\mathbf A_t^{\pi}$. It uses a cosine schedule, a minimum SNR weighting threshold of 5, 50 diffusion steps during training, and 16 DDIM steps for sampling. Its checkpoint is selected using sampled action MAE. Rectified flow matching~\cite{lipman2023flow} trains the same UNet to predict $\mathbf A_t^{\pi}-\boldsymbol\varepsilon$ and generates a chunk with 20 Euler steps. Its checkpoint is selected using flow MSE. The two methods therefore apply different training objectives to the same network rather than using separate architectures.

Candidate chunks are scored by a causal action Transformer, denoted $W_{\mathrm T}$. Like AC-GRU-WM, it predicts step $\tau$ using the history $\mathbf Y_t$ and only the commands available up to that step. The Transformer has width 256, eight attention heads, three encoder layers, and four decoder layers, with layer normalisation before each block. Training uses dropout $0.1$, a learning rate of $3{\times}10^{-4}$, at most 40 epochs, and extra weight on retained liquid. The model predicts changes from the latest state and accumulates them through the rollout. Checkpoints are selected using mean water MAE over horizons from $0.16$ to $1.28\,\mathrm{s}$. Test error increases from $2.43\,\mathrm{g}$ on the smallest sponge to $4.52\,\mathrm{g}$ on the largest. We use $W_{\mathrm T}$ for ranking because the recurrent model does not resolve the action effect reliably on the largest sponge.

For each test window, we draw $K=8$ candidate chunks and select
\begin{subequations}\label{eq:rank}
\begin{align}
&\mathbf A_t^{\pi,\star} = \arg\min_{k}J\!\left(\hat{\mathbf y}^{(k)},\mathbf A_t^{\pi,(k)},g_t\right), \label{eq:rank_selection}\\
&J = \lambda_M J_M+\lambda_F J_F+\lambda_S J_S+\lambda_U J_U. \label{eq:rank_score}
\end{align}
\end{subequations}
Here $J_M$ is the terminal liquid mass error predicted by $W_{\mathrm T}$, $J_F$ penalises peak force, $J_S$ is the predicted slip probability, and $J_U$ penalises action jerk. Their weights are $\lambda_M=1$, $\lambda_F=10^{-4}$, $\lambda_S=0.05$, and $\lambda_U=0.1$. We report the force MAE of one policy sample and the predicted terminal liquid mass error for the collected reference chunk, the first policy sample, and the selected chunk $\mathbf A_t^{\pi,\star}$.

\section{Experiments}
\label{sec:experiments}

Experiments use the robot augmented SPH environment to perform scripted grasp, lift, and squeeze trials on a sponge filled with water. We test three sponge sizes, $8{\times}8{\times}18$, $10{\times}10{\times}20$, and $12{\times}12{\times}22\,\mathrm{cm}$, at porosities of $0.5$, $0.6$, and $0.7$. The force used to hold the sponge before lifting and the force used to squeeze it afterwards are varied factorially. Young's modulus is $250\,\mathrm{kPa}$, the static and kinetic friction coefficients are $0.6$ and $0.5$, and the recovery gain is $\beta=0.2$. Only trajectories that complete a stable lift and squeeze are used to train the world models and policies.

The collected dataset contains 594 runs over the complete parameter grid. Of these, 410 complete a stable squeeze after lifting and form the learning dataset; the remaining runs are retained for analysing failure and physical behaviour. The stable runs cover all three sponge sizes. For every combination of size, porosity, and holding force, the six squeeze forces form a matched response family. The state before each action is checked and adjusted rather than assumed to be identical.

\begin{figure}[t]
\centering
\includegraphics[width=0.8\linewidth]{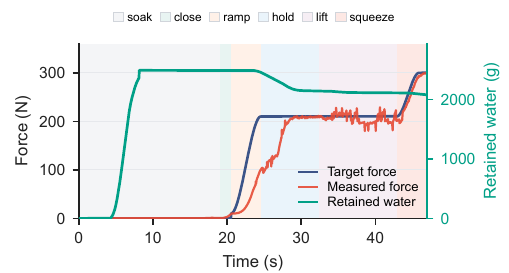}
\caption{One completed run for the medium sponge at $\phi=0.7$, using $210\,\mathrm{N}$ to hold the sponge before lifting and $300\,\mathrm{N}$ to squeeze it afterwards. The left axis shows target and measured gripper force, while the right axis shows retained water. Most liquid leaves as the measured force approaches the holding target. Lifting changes the retained water only modestly, and the later squeeze removes a smaller additional amount. Water is shown as a $2\,\mathrm{s}$ trailing average.}
\label{fig:single_run}
\end{figure}

\begin{figure}[t]
\centering
\includegraphics[width=0.75\linewidth]{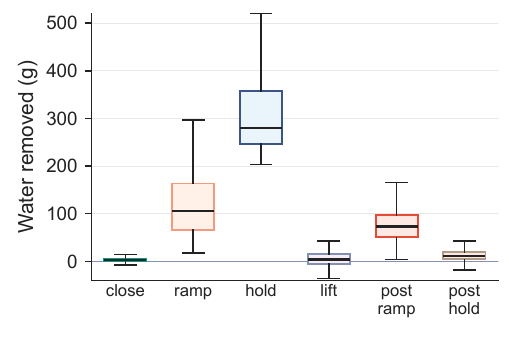}
\caption{Water removed during each controller phase over all runs that reach that phase. The initial force ramp and hold dominate transport, while lifting has the smallest median effect. The post-ramp and post-hold phases constitute the later squeeze stage, which removes part of the remaining liquid. The distributions show that clamping and holding, rather than lifting, account for most transport through the porous material.}
\label{fig:water}
\end{figure}

Trajectories are resampled on a global $20\,\mathrm{ms}$ time grid. Training windows use stride one, a history of $L=16$, and a rollout of $H=64$, with forecasts evaluated from $0.16$ to $1.28\,\mathrm{s}$. Windows exclude the duplicated \textsc{waiting} phase and the initial velocity command, and they never cross run boundaries. Within each sponge size, all force variants that share the same porosity and holding force remain in the same partition. Using seed 0, the grouped split assigns $70\%$ to training, $10\%$ to validation, and $20\%$ to testing. Preprocessing and checkpoint selection use only the training and validation partitions. Results are reported separately for each sponge size. We first present physical behaviour, causal prediction, and policy ranking as the main results, followed by ablations of action information, state representation, access to the complete future chunk, and policy training objective.

\subsection{Evaluation Metrics}

Forecasting is evaluated using rollout MAE for retained water in grams and measured force in newtons after conversion back to physical units. Because the four friction regimes are imbalanced, we report balanced accuracy rather than raw accuracy. Contact balanced accuracy is $1.0$ at $0.64\,\mathrm{s}$ for every method because these windows already contain contact, so it cannot distinguish the models. Policy force MAE and the mean absolute force change per step measure how closely one sample follows the collected reference command. The scorer's water error is $\hat E_M^{W_{\mathrm T}}=|\hat M_{t+H_\pi}^{L}-g_t|$, measured in grams at the end of the rollout. All results are computed separately for each sponge size.

\begin{figure*}[t]
\centering
\includegraphics[width=0.85\textwidth]{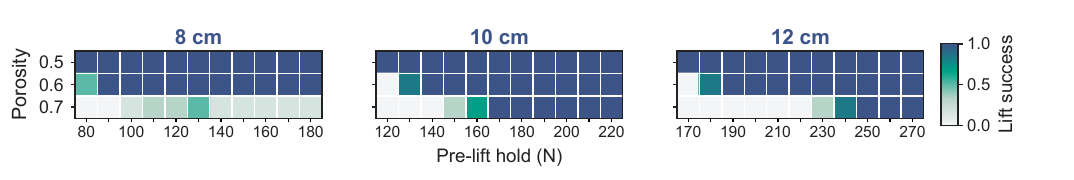}
\caption{Lift success over porosity and the force used to hold the sponge before lifting. Each cell combines the six later squeeze settings. All conditions with $\phi=0.5$ lift successfully. Higher porosity requires a stronger holding force and remains difficult for the small and large sponges.}
\label{fig:lift_surface}
\end{figure*}

\begin{table*}[t]
\caption{Causal world model prediction on the collected test split at $0.64\,\mathrm{s}$. We report water MAE in grams, force MAE in newtons, and balanced accuracy over four friction classes. The first three rows form the main comparison, and bold marks the best of these models. The final row is the action input ablation discussed in Sec.~\ref{sec:ablations}.}
\label{tab:forecast}
\centering
\scriptsize
\setlength{\tabcolsep}{3.2pt}
\begin{tabular}{lccccccccc}
\toprule
& \multicolumn{3}{c}{$8\,\mathrm{cm}$} &
\multicolumn{3}{c}{$10\,\mathrm{cm}$} &
\multicolumn{3}{c}{$12\,\mathrm{cm}$}\\
\cmidrule(lr){2-4}\cmidrule(lr){5-7}\cmidrule(lr){8-10}
Method & water ($\mathrm{g}$) & force ($\mathrm{N}$) & fric.  & water ($\mathrm{g}$) & force ($\mathrm{N}$) & fric.  & water ($\mathrm{g}$) & force ($\mathrm{N}$) & fric. \\
\midrule
Baseline & 12.23 & 7.93 & 0.57 & 16.71 & 10.01 & 0.61 & 23.31 & \textbf{10.94} & 0.58\\
State-GRU-WM & 9.42 & 9.62 & 0.61 & 9.02 & 6.77 & 0.65 & \textbf{17.03} & 12.98 & \textbf{0.68}\\
AC-GRU-WM & \textbf{4.32} & \textbf{3.78} & \textbf{0.72} & \textbf{6.59} & \textbf{4.74} & \textbf{0.79} & 17.53 & 11.04 & 0.61\\
\midrule
\multicolumn{10}{l}{\emph{Action input ablation}}\\
AC-GRU-WM, shuffled $\mathbf A$ & 15.28 & 15.41 & 0.55 & 15.68 & 14.67 & 0.60 & 18.49 & 11.27 & 0.61\\
\bottomrule
\end{tabular}
\end{table*}

\section{Results}
\label{sec:results}

\subsection{Main Results}

\paragraph{Physical behaviour of porous RMI}

The 410 trajectories used for learning complete the grasp, lift, and squeeze sequence across all three sponge sizes. Most liquid leaves during the initial force ramp and hold, lifting changes retained water comparatively little, and the later squeeze removes part of what remains (Figs.~\ref{fig:single_run} and~\ref{fig:water}). The main transport process is therefore pore collapse during the initial clamp. Later actions mainly affect the remaining liquid and grasp stability, which are the outcomes predicted by the learned models.

The factorial analysis also reveals a grasp boundary (Fig.~\ref{fig:lift_surface}). Every sponge lifts successfully at $\phi=0.5$, but high porosity makes lifting much less reliable. At $\phi=0.7$, success ranges from $20\%$ to $64\%$, with the medium sponge proving the most robust. Within groups matched by size, porosity, and holding force, each additional newton of squeeze force is associated with $0.669\,\mathrm{g}$ of additional water removal. The clustered 95\% confidence interval ranges from $0.610$ to $0.728\,\mathrm{g}$. Because the amount of water before the action still varies across force settings, this result is an association within the factorial design rather than an isolated causal force sweep.

\paragraph{Causal world model prediction}

The first three rows of Table~\ref{tab:forecast} compare AC-GRU-WM with the Baseline and the matched recurrent model without actions. On the small and medium sponges, AC-GRU-WM reduces water error by more than $60\%$ relative to the Baseline and also improves friction classification. It outperforms State-GRU-WM as well, showing that recent state history alone is insufficient in these conditions. On the largest sponge, however, action conditioning provides no clear benefit. We retain this negative result because it limits the claim: the causal world model predicts the effect of a command only when that effect is identifiable in the collected trajectories.

\paragraph{Policy generation and world model selection}

The main policy result asks whether $W_{\mathrm T}$ can select a better command from eight proposals. Before selection, the first proposal from either policy remains close to the collected reference command for every sponge size. Table~\ref{tab:ranking} shows that ranking consistently lowers predicted terminal water error. Flow matching reduces the error by roughly half, and Diffusion Policy shows the same overall trend. The combined score $J$ also improves for every size. Agreement across both proposal methods shows that the policy produces useful variation and that the world model can identify commands with a better predicted phase outcome.

\begin{table}[t]
\caption{Predicted terminal water error in grams for the original controller action, the first policy sample, and the action selected from eight samples by Transformer $W_{\mathrm T}$.}
\label{tab:ranking}
\centering
\scriptsize
\setlength{\tabcolsep}{2.2pt}
\begin{tabular}{lccc}
\toprule
Method & $8\,\mathrm{cm}$ ($\mathrm{g}$) & $10\,\mathrm{cm}$ ($\mathrm{g}$) & $12\,\mathrm{cm}$ ($\mathrm{g}$)\\
\midrule
Original controller action & 2.56 & 4.23 & 4.68\\
Diffusion policy, first sample & 2.79 & 4.30 & 4.94\\
Diffusion policy $+\ W_{\mathrm T}$ & 1.24 & 2.57 & 2.82\\
Flow matching, first sample & 2.61 & 4.28 & 4.86\\
Flow matching $+\ W_{\mathrm T}$ & 1.11 & 2.24 & 2.45\\
\bottomrule
\end{tabular}
\end{table}

\subsection{Ablation Studies}
\label{sec:ablations}

\paragraph{Action information}

The last row of Table~\ref{tab:forecast} keeps the AC-GRU-WM architecture unchanged but assigns each state history a future command from another trajectory. This preserves the overall command distribution while breaking the relationship between a command and its outcome. Water and friction performance deteriorate sharply on the small and medium sponges, confirming that the model uses the command rather than only following recent drainage. Shuffling has little effect on the largest sponge, which agrees with the negative result in the main comparison. State-GRU-WM provides a second control by removing the command entirely.

\paragraph{State representation}

Table~\ref{tab:ablation} progressively adds gripper loading, contact, and residual slip to the liquid and solid phase variables. The complete state, which has 23 dimensions, is best for the small and medium sponges, where the command is identifiable. The pattern reverses on the largest sponge: the phase variables alone outperform the complete state by a wide margin. The interaction variables therefore help when the command effect can be resolved, while the phase variables provide the most reliable signal in the unresolved condition.

\begin{table}[t]
\caption{Ablation of the state used by AC-GRU-WM at $0.64\,\mathrm{s}$, reported as water MAE in grams. LS contains liquid and solid phase variables, G adds gripper loading, C adds contact, and ALL adds residual slip.}
\label{tab:ablation}
\centering
\scriptsize
\setlength{\tabcolsep}{4pt}
\begin{tabular}{lccc}
\toprule
State & $8\,\mathrm{cm}$ ($\mathrm{g}$) & $10\,\mathrm{cm}$ ($\mathrm{g}$) & $12\,\mathrm{cm}$ ($\mathrm{g}$)\\
\midrule
LS & 9.11 & 10.85 & \textbf{8.41}\\
LS$+$G & 8.68 & 11.70 & 14.63\\
LS$+$G$+$C & 7.36 & 13.32 & 14.74\\
ALL & \textbf{4.32} & \textbf{6.59} & 17.53\\
\bottomrule
\end{tabular}
\end{table}

\paragraph{Access to the complete action chunk}

Table~\ref{tab:oneshot} tests models that receive the complete future action chunk at once. AC-MLP-WM keeps water error below $8\,\mathrm{g}$ for every sponge size. This accuracy depends strongly on the future command because shuffling the chunk raises the error to between $17$ and $43\,\mathrm{g}$. Linear-ARX predicts force well only on the smallest sponge, and its error grows sharply with size. Access to later commands therefore helps interpolate the collected trajectories, but this diagnostic is not causal and does not replace the world model used for robot decisions.

\begin{table}[t]
\caption{Diagnostic in which the complete future action chunk is visible at every predicted step. Water and force MAE are reported at $0.64\,\mathrm{s}$ in grams and newtons. Bold marks the best result within this diagnostic.}
\label{tab:oneshot}
\centering
\scriptsize
\setlength{\tabcolsep}{2.2pt}
\begin{tabular}{lcccccc}
\toprule
Method & \multicolumn{2}{c}{$8\,\mathrm{cm}$} ($\mathrm{g|N}$) &
\multicolumn{2}{c}{$10\,\mathrm{cm}$} ($\mathrm{g|N}$) &
\multicolumn{2}{c}{$12\,\mathrm{cm}$} ($\mathrm{g|N}$)\\
& water  & force & water & force & water & force\\
\midrule
Linear-ARX & 4.29 & \textbf{1.77} & 9.53 & 13.48 & 9.67 & 25.77\\
AC-MLP-WM & \textbf{3.74} & 2.72 & \textbf{5.75} & \textbf{3.96} & \textbf{7.44} & \textbf{5.87}\\
AC-MLP-WM, shuffled $\mathbf A$ & 17.47 & 13.41 & 26.15 & 19.60 & 42.87 & 26.02\\
\bottomrule
\end{tabular}
\end{table}

\paragraph{Policy training objective}

Table~\ref{tab:imitation} changes only the objective used to train the shared UNet. Diffusion Policy and flow matching both reproduce the collected force commands with roughly $5$ to $6\,\mathrm{N}$ MAE, and both generate smooth changes in force. They differ by less than one newton on every sponge size. The main ranking result therefore does not depend on one particular generative objective.

\begin{table}[t]
\caption{Ablation of the policy training objective on a shared temporal UNet. We report force MAE and the mean force change per step for one sample, with both quantities measured in newtons. Each sponge size contributes several thousand test windows.}
\label{tab:imitation}
\centering
\scriptsize
\setlength{\tabcolsep}{2.0pt}
\resizebox{\columnwidth}{!}{%
\begin{tabular}{lcccccc}
\toprule
Method & \multicolumn{2}{c}{$8\,\mathrm{cm}$ ($\mathrm{N}$)}  &
\multicolumn{2}{c}{$10\,\mathrm{cm}\, (\mathrm{N})$} &
\multicolumn{2}{c}{$12\,\mathrm{cm}\, (\mathrm{N})$} \\
& MAE$_F$ & smooth$_F$ & MAE$_F$ & smooth$_F$ & MAE$_F$ & smooth$_F$\\
\midrule
Diffusion policy & 5.23 & 0.31 & 5.87 & 0.27 & 4.79 & 0.34\\
Flow matching & 5.11 & 0.35 & 6.26 & 0.31 & 5.37 & 0.37\\
\bottomrule
\end{tabular}
}
\end{table}

\section{Conclusion}

Most manipulation methods model rigid or deformable solids, while methods for liquid manipulation usually keep the liquid outside the solid. We introduced Robotic Multiphase Interaction for a different case: liquid occupies a porous solid and evolves through mechanical coupling with its deforming skeleton. A robot interaction layer with contact memory, hybrid control, and a stability gate turns porous flow SPH into a robot-interactive environment. We define compact state and action representations that connect the underlying particle dynamics to robot learning. Based on this representation, an action-conditioned world model predicts how the coupled liquid and solid state changes under robot commands, while a generative policy proposes actions for goals such as controlling the amount of retained water. On the small and medium sponges, the causal world model reduces water prediction error by more than $60\%$ relative to the Baseline. Ranking generated actions with a causal Transformer also reduces the predicted terminal water error by roughly half. These results show that RMI shifts the manipulation question from how an action moves or deforms a solid to how it changes the interaction between liquid and the porous solid that contains it. Future work will transfer the world model and policy to physical sponge manipulation and study their generalisation across porous materials.

\balance
\bibliographystyle{IEEEtran}
\bibliography{rmi_icra_revised}

\clearpage

\clearpage

\setcounter{section}{0}
\setcounter{subsection}{0}
\setcounter{equation}{0}
\setcounter{figure}{0}
\setcounter{table}{0}

\renewcommand{\thesection}{S\arabic{section}}
\renewcommand{\thesubsection}{\thesection.\arabic{subsection}}
\renewcommand{\theequation}{S\arabic{equation}}
\renewcommand{\thefigure}{S\arabic{figure}}
\renewcommand{\thetable}{S\arabic{table}}

\begin{center}
{\LARGE\bfseries Supplementary Material}\\[0.8em]
\end{center}

\vspace{1em}

This supplement provides the implementation details that are omitted from the main paper. It covers the porous flow quantities used to construct the interaction state, the particle level contact model, the quantities used by the lift gate, the exact learning losses, the experimental split, and one additional policy selection result.

\section{Porous Flow Details}
\label{supp:sec:physics}

The environment uses Implicit Incompressible Porous Flow with Smoothed Particle Hydrodynamics (SPH)~\cite{boettcher2025porous}. Liquid and solid particles overlap spatially but remain distinct phases. Indices $i,j$ denote liquid particles, while $s,r$ denote solid particles. The undeformed porosity is $\phi\in(0,1)$.

For the reference sampling volume $V_s^0$ and solid material rest density $\rho_s^0$, the mass of solid particle $s$ is
\begin{equation}
m_s=(1-\phi)\rho_s^0V_s^0.
\label{supp:eq:solid_mass}
\end{equation}
Its current density is estimated from neighbouring solid particles as
\begin{equation}
\rho_s=(1-\phi)\rho_s^0\sum_{r\in\mathcal N_s^S}V_r^0\mathcal K_{sr}.
\label{supp:eq:solid_density}
\end{equation}
The liquid density includes the local volume occupied by the solid skeleton:
\begin{equation}
\begin{aligned}
\hat\rho_i={}&\rho_i^0\sum_{j\in\mathcal N_i^L}V_j^0\mathcal K_{ij}\\
&+\rho_i^0\sum_{r\in\mathcal N_i^S}(1-\phi)V_r^0\mathcal K_{ir}.
\end{aligned}
\label{supp:eq:liquid_density}
\end{equation}
Here $\mathcal K_{ab}=\mathcal K(\|\mathbf x_a-\mathbf x_b\|,h)$ is the SPH kernel with smoothing length $h$. Equations~\eqref{supp:eq:solid_density} and~\eqref{supp:eq:liquid_density} allow the two particle sets to sample the same region without assigning the entire volume to either phase. The saturation in Eq.~\eqref{eq:saturation} then relates local liquid content to the current volume of the porous skeleton.

\section{Robot Contact Details}
\label{supp:sec:interaction}

\subsection{Contact Memory}

Consider solid particle $s$ in contact with a gripper. Before gripper friction is applied, the porous solver predicts
\begin{equation}
\bfv_s^{*}=\bfv_s^{n}+\Delta t(\bfa_s^{\mathrm{np}}+\bfa_s^{\mathrm{p}}).
\label{supp:eq:free_velocity}
\end{equation}
Here $\bfa_s^{\mathrm{np}}$ contains nonpressure acceleration and $\bfa_s^{\mathrm{p}}$ contains pressure acceleration. For gripper centre $\bfp_b$, orientation $\mathbf R_b$, linear velocity $\bfv_b$, and angular velocity $\bfomega_b$, the velocity at contact point $\bfp_{c,s}$ is
\begin{equation}
\bfv_b(\bfp_{c,s})=\bfv_b+\bfomega_b\times(\bfp_{c,s}-\bfp_b).
\label{supp:eq:gripper_velocity}
\end{equation}
The tangential velocity before friction is
\begin{equation}
\vtan_s=(\mathbf I-\bfn_s\bfn_s^\top)\left[\bfv_s^{*}-\bfv_b(\bfp_{c,s})\right],
\label{supp:eq:tangential_velocity}
\end{equation}
where $\bfn_s$ is the contact normal.

When the contact sticks, its position is stored in the gripper frame. The stored point and its tangential displacement are
\begin{subequations}\label{supp:eq:anchor}
\begin{align}
&\bfq_{a,s}=\mathbf R_b^\top(\bfp_{c,s}-\bfp_b), \label{supp:eq:anchor_local}\\
&\bfp_{a,s}=\bfp_b+\mathbf R_b\bfq_{a,s}, \label{supp:eq:anchor_world}\\
&\mathbf e_s=(\mathbf I-\bfn_s\bfn_s^\top)(\bfp_{a,s}-\bfp_{c,s}). \label{supp:eq:anchor_error}
\end{align}
\end{subequations}
Equation~\eqref{eq:friction_demand} uses $\mathbf e_s$ to compute the recovery demand. A sticking contact keeps its stored point while the demand remains within the static Coulomb bound. When the bound is exceeded, the contact switches to kinetic slip and discards the stored point. A new point is created when sticking is established again.

After applying the friction force $\mathbf F_{f,s}$ from Eq.~\eqref{eq:friction_applied}, the particle state is updated by
\begin{subequations}\label{supp:eq:contact_update}
\begin{align}
&\bfv_s^{n+1}=\bfv_s^{*}+\frac{\Delta t}{m_s}\mathbf F_{f,s}, \label{supp:eq:velocity_update}\\
&\bfp_s^{n+1}=\bfp_s^{n}+\Delta t\,\bfv_s^{n+1}. \label{supp:eq:position_update}
\end{align}
\end{subequations}
The corrected tangential velocity is
\begin{equation}
\begin{aligned}
\mathbf v_s^{\mathrm{tan,corr}}={}&(\mathbf I-\bfn_s\bfn_s^\top)\\
&\left[\bfv_s^{n+1}-\bfv_b(\bfp_{c,s})\right].
\end{aligned}
\label{supp:eq:corrected_tangential_velocity}
\end{equation}
Residual slip is the mean magnitude of this velocity over the valid contact set:
\begin{equation}
\vres_t=\frac{1}{|\mathcal C_t|}\sum_{s\in\mathcal C_t}\|\mathbf v_{s,t}^{\mathrm{tan,corr}}\|.
\label{supp:eq:residual_slip}
\end{equation}
The average uses magnitudes so that motion in opposite directions cannot cancel.

\subsection{Lift Gate Quantities}

The lift gate uses exponentially filtered left and right gripper loads:
\begin{equation}
\tilde F_{b,t}=(1-\eta_F)\tilde F_{b,t-1}+\eta_F F_{b,t}.
\label{supp:eq:force_filter}
\end{equation}
Here $b\in\{L,R\}$ identifies the gripper side and $\eta_F\in(0,1]$ is the filter coefficient. For the set of solid particles $\mathcal S$ and intended lift direction $\bfa_{\mathrm{lift}}$, the centre of mass (COM) velocity and its component along the lift direction are
\begin{subequations}\label{supp:eq:com_motion}
\begin{align}
&\mathbf V_{\mathrm{COM},t}=\frac{\sum_{s\in\mathcal S}m_s\bfv_{s,t}}{\sum_{s\in\mathcal S}m_s}, \label{supp:eq:com_velocity}\\
&v_{\mathrm{COM},t}=|\mathbf V_{\mathrm{COM},t}\cdot\bfa_{\mathrm{lift}}|. \label{supp:eq:com_lift_speed}
\end{align}
\end{subequations}
The filtered loads, COM speed, and residual slip are evaluated using the conditions in Eq.~\eqref{eq:lift_gate}. All conditions must remain valid for $K_{\mathrm{stable}}$ consecutive simulator steps before both grippers receive the same lift displacement.

\section{Learning Details}
\label{supp:sec:learning}

The predictor is the Action Conditioned Gated Recurrent Unit World Model (AC-GRU-WM). It uses the state ordering in Eq.~\eqref{eq:interaction_state}, with 18 continuous values, one contact indicator, and a four-class friction vector. Continuous state and action channels are standardised using statistics from the training partition:
\begin{subequations}\label{supp:eq:normalisation}
\begin{align}
&\bar z_{t,d}=\frac{z_{t,d}-\mu_d}{s_d+\varepsilon_N}, \label{supp:eq:state_normalisation}\\
&\bar u_{t,j}=\frac{u_{t,j}-\mu_j^u}{s_j^u+\varepsilon_A}. \label{supp:eq:action_normalisation}
\end{align}
\end{subequations}
Here $\mu$ and $s$ are the corresponding training means and standard deviations, while $\varepsilon_N$ and $\varepsilon_A$ prevent division by zero. Binary and categorical values are not standardised. The validity mask removes only undefined targets. Physical zeros, including zero contact area and zero applied friction when contact is absent, remain valid.

The continuous loss $\mathcal L_{\mathrm{cont}}$ is defined in Eq.~\eqref{eq:continuous_loss}. The classification losses and the complete objective are
\begin{subequations}\label{supp:eq:complete_loss}
\begin{align}
&\mathcal L_c=\sum_{\tau=1}^{H}\gamma^{\tau-1}\operatorname{BCE}(\hat p_{t+\tau}^{c},c_{t+\tau}), \label{supp:eq:contact_loss}\\
&\mathcal L_f=\sum_{\tau=1}^{H}\gamma^{\tau-1}\operatorname{CE}(\hat{\mathbf P}_{t+\tau}^{\mathrm{fric}},\mathbf r_{t+\tau}^{\mathrm{fric}}), \label{supp:eq:friction_loss}\\
&\mathcal L_{\mathrm{WM}}=\mathcal L_{\mathrm{cont}}+\lambda_c\mathcal L_c+\lambda_f\mathcal L_f. \label{supp:eq:world_model_loss}
\end{align}
\end{subequations}
We use $\gamma=0.98$ and $\lambda_c=\lambda_f=0.2$. Teacher forcing is reduced linearly from one to zero during the first 20 epochs. Training stops after at most 40 epochs, with patience of eight epochs. The checkpoint is selected using free rollout retained water mean absolute error (MAE) at $H=64$ on the validation partition.

The causal Transformer scorer uses additional weight eight on the retained water channel. Its checkpoint is selected using the mean validation water MAE at horizons of $0.16$, $0.32$, $0.64$, and $1.28\,\mathrm{s}$. Test trajectories are not used for normalisation, training, or checkpoint selection.

\section{Dataset Details}
\label{supp:sec:data}

The dataset contains 594 trajectories across three sponge sizes and three porosities. The learning dataset contains the 410 trajectories that complete both lifting and squeezing with a stable grasp. It includes 110 trajectories for the small sponge, 157 for the medium sponge, and 143 for the large sponge. The other trajectories are used only to analyse physical behaviour and failure.

All trajectories are resampled on the global $20\,\mathrm{ms}$ grid described in Sec.~\ref{sec:experiments}. Windows never cross trajectory boundaries. Conditions that share sponge size, porosity, and holding force remain in the same training, validation, or test partition, so later squeeze variants from the same response family cannot appear on both sides of the split. Normalisation and checkpoint selection use only the training and validation partitions.

The dataset does not vary target force and direct velocity independently, and the trials do not begin from identical particle states. Initial liquid mass is also not varied independently of the soaking procedure. These limitations make some action effects difficult to identify, particularly for the largest sponge, and motivate the separate causal and lookahead evaluations in the main paper.

\section{Policy Selection Cost}
\label{supp:sec:policy_cost}

The main paper reports terminal water error for the Diffusion Policy U-Net (DP-UNet) and Flow Matching U-Net (FM-UNet). Table~\ref{supp:tab:policy_cost} reports the corresponding combined cost $J$ from Eq.~\eqref{eq:rank_score}. The original controller action is the command sequence used to generate the corresponding test trajectory and is not assumed to be optimal. For each policy, $W_{\mathrm T}$ selects one action from eight candidates.

\begin{table}[t]
\caption{Combined predicted cost $J$ for the original and selected actions.}
\label{supp:tab:policy_cost}
\centering
\scriptsize
\setlength{\tabcolsep}{3pt}
\begin{tabular}{lccc}
\toprule
Action source & $8\,\mathrm{cm}$ & $10\,\mathrm{cm}$ & $12\,\mathrm{cm}$\\
\midrule
Original controller action & 3.61 & 5.22 & 6.00\\
DP-UNet $+\ W_{\mathrm T}$ & 2.61 & 3.69 & 4.37\\
FM-UNet $+\ W_{\mathrm T}$ & 2.51 & 3.43 & 4.06\\
\bottomrule
\end{tabular}
\end{table}

The selected actions reduce the combined predicted cost for every sponge size under both policy objectives. This result supports the use of the world model to choose among policy proposals without repeating the terminal water results reported in the main paper.

\end{document}